# INTRUSION DETECTION: MACHINE LEARNING BASELINE CALCULATIONS FOR IMAGE CLASSIFICATION

Erik Larsen, Korey MacVittie, John Lilly

PeopleTec, Inc., 4901 Corporate Drive. NW, Huntsville, AL, USA
erik.larsen@peopletec.com

## ABSTRACT

*Cyber security can be enhanced through application of machine learning by recasting network attack data into an image format, then applying supervised computer vision and other machine learning techniques to detect malicious specimens. Exploratory data analysis reveals little correlation and few distinguishing characteristics between the ten classes of malware used in this study. A general model comparison demonstrates that the most promising candidates for consideration are Light Gradient Boosting Machine, Random Forest Classifier, and Extra Trees Classifier. Convolutional networks fail to deliver their outstanding classification ability, being surpassed by a simple, fully connected architecture. Most tests fail to break 80% categorical accuracy and present low F1 scores, indicating more sophisticated approaches (e.g., bootstrapping, random samples, and feature selection) may be required to maximize performance.*

## 1. INTRODUCTION

The UNSW-NB15 network attack dataset [1] has been recast into an image dataset [2] and stored on Kaggle [3]. This results in 257,673 total thumbnail images divided into families for each of 10 classes of intruder: *generic*, *normal*, *exploits*, *backdoor*, *fuzzers*, *reconnaissance*, *analysis*, *shellcode*, *dos*, and *worms*, as seen in Figure 1.

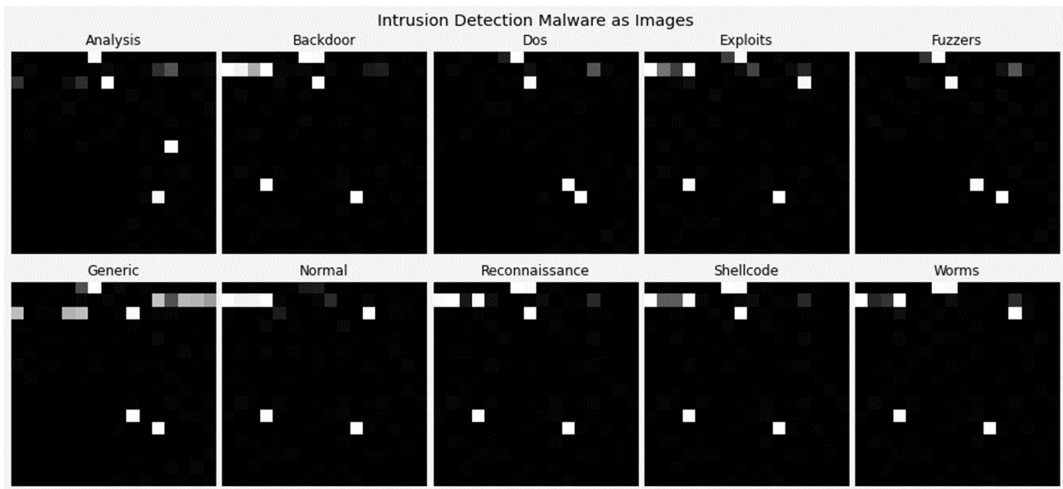

Figure 1. Random thumbnail examples

175,341 of the total examples are labelled as the test set, leaving 82,332 as training. These training and test set sizes are peculiar: while it is beneficial for the training set to be very large, here we see a larger test set causing an alarming test/train ratio of 2.13. Shuffling the two sets together, then re-splitting into training, validation, and test sets with respective ratios of 0.90 : 0.05 : 0.05 generated the results contained herein.

Intrusion detection grayscale images are all arrays of shape (16, 16, 1) [3]. This is convenient because it requires no additional reshaping for ingestion into a convolutional neural network (CNN). However, Figure 2 shows that support for each type of malware is not uniform across the population. Some — e.g., worms with only 177 thumbnails total — are disproportionately represented. These ratios are replicated through stratification when re-splitting data into train and test sets.

Excellent scores (i.e., greater than 0.90 in desired metrics) are possible even if targets with low support are all misidentified by investigating individual performances instead of the gestalt. For example, some tests yield no correct classifications for the analysis, backdoor, and worm species. If persistent classification errors develop it may be necessary to acquire more examples of under-represented classes or create synthetic data to increase their numbers.

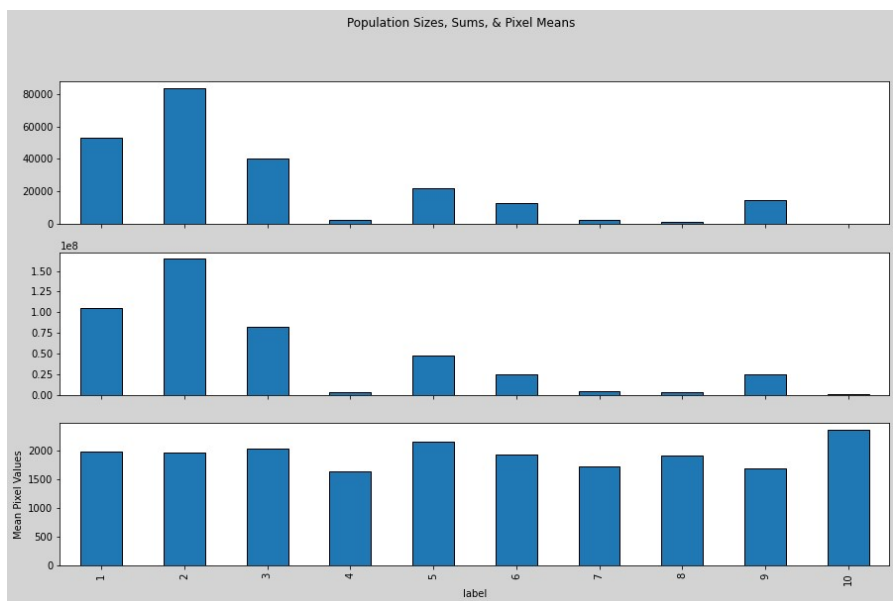

Figure 2. Population disparity and pixel sum similarity

## 1.1. Methods

This research uses both Kaggle and Google Colab (Jupyter) notebooks with the Overhead-MNIST dataset found at Kaggle.com [2-6], and individual Python IDEs. ML libraries used include Pycaret, UMAP, sci-kit-learn, and TensorFlow's Keras API (2.4.0). Training and evaluation occur on a total of 23 classifiers [1, 8-12]. The given test set remains unseen by the algorithm and is used for the final model scoring independent from Pycaret.

## 1.2. Data Preparation

Raw data augmentation is not performed on the sparse thumbnail arrays. Detailed image processing is not evaluated. The presence of dissimilar class sizes necessitates stratification when splitting the training set into train and validation sets; we use 5% of the training data for validation. Final results are based on predictions made from unseen test data.

The new training set consists of 180,371 examples from the ten different classes. Each has 258 features, the first 256 of which are pixel values. The remaining two are "label" and "class". The label gives a numerical value for the class, so that either can be used when appropriate. All data is of type int64 except 'class,' which is a string object. Because class numbers start with "1" yet Python indexing begins with "0," 1 is subtracted from the labels before transforming into categorical vectors.

## 2. EXPLORATORY INSIGHTS

Visualizations such as Figure 3 demonstrate these classes are not as distinguishable as those found in the Virus-MNIST data set [11, 12]. Even descriptive statistics tease out only slight differences between a few classes. The classes share relatively close average pixel values, as seen in Figure 4, whereby class ten clearly has the greatest of these. The most pronounced feature is the maximum pixel value, which tends to be very low in the bottom and lower right corner of the images (see Figure 5). There are no missing values. The lack of distinguishing physical features (e.g., textures and horizontal edges) in these images could make an image classification approach less viable.

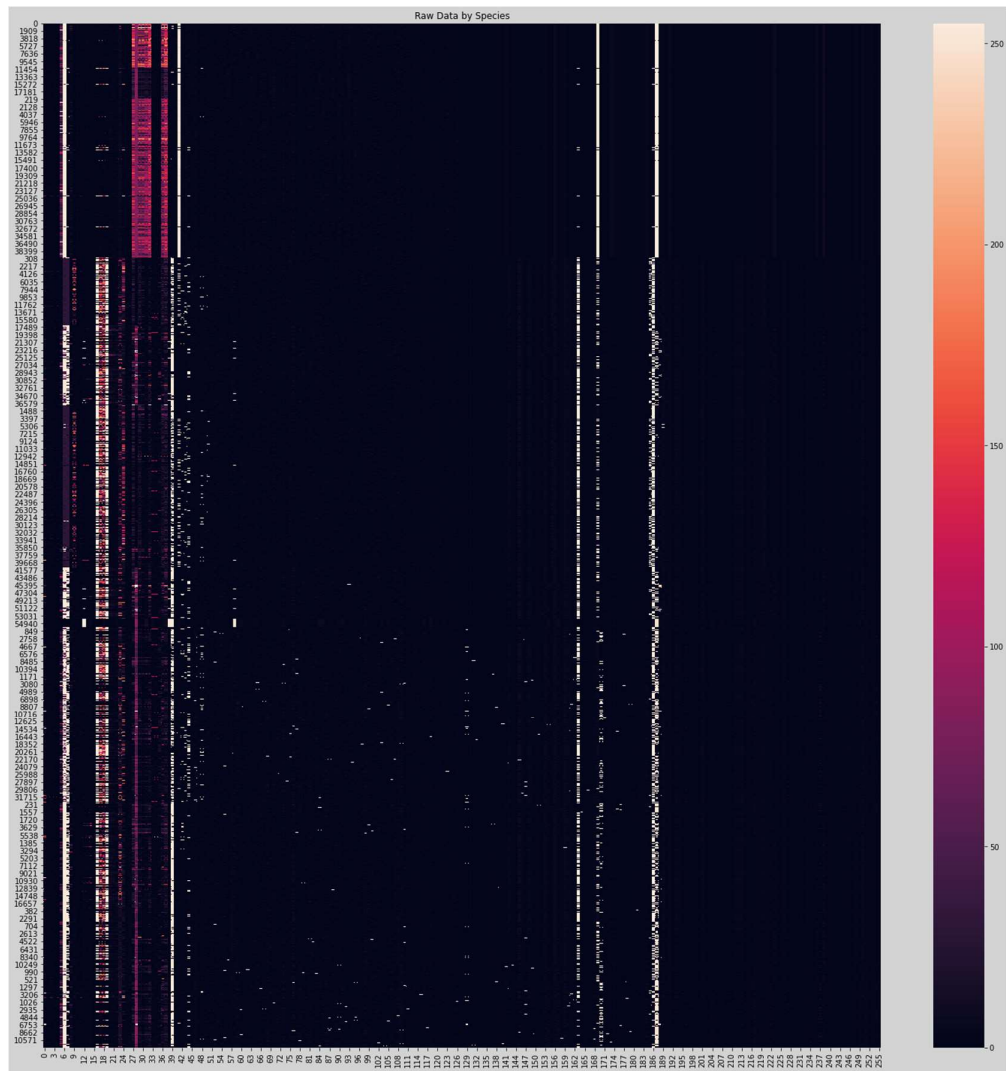

Figure 3. Heatmap of pixel values grouped by class

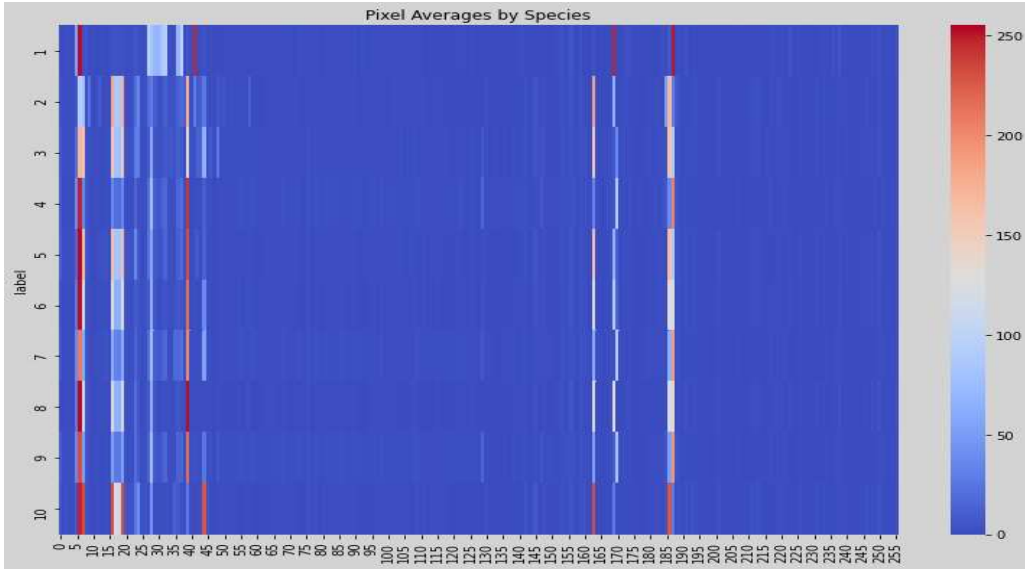

Figure 4. Averages appear similar across labels, but some differentiation is apparent

The arrays' lack of discerning characteristics makes categorizing these images difficult for convolutional methods, emphasizing the importance of carefully considering the validity of any model chosen for deployment. A large mismatch between target image populations can yield the semblance of high accuracy scores yet exhibit poor individual performance and significantly lower F1 scores.

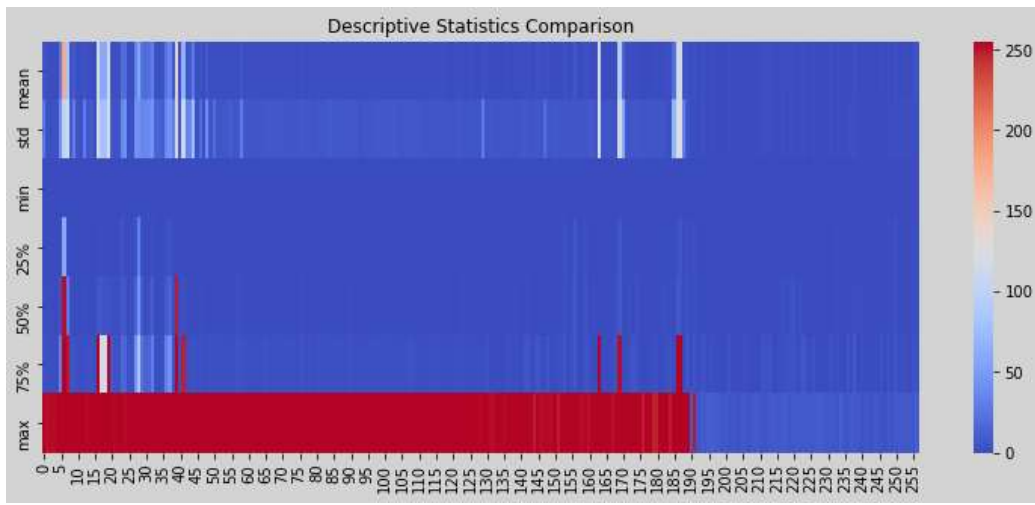

Figure 5. Heatmap reveals a steep drop off in maximum value for pixels in the lower right along with a heavy skew towards low values across the first three quartiles

## 3. MODEL COMPARISON RESULTS

### 3.1. Pycaret

PyCaret [13] is an open-source library for quickly comparing multiple common machine learning algorithms on given data. Light Gradient Boosting Machine (lightgbm), Random Forest Classifier (rf), and Extra Trees Classifier (et) are the top three scoring algorithms with accuracies of 0.7787,

0.7743 and 0.7709, respectively. Other model scores are reported in Table 1. The vast number of examples necessitates using only two cross-validation folds to avoid timeout. Curiously, during a 3-fold run the Logistic Regression (LR) classifier scored near 100%, though the exceedingly long duration was not recorded before timeout and this was considered anomalous. With two folds, LR only achieved 66%, ranking 7th overall.

The top two methods have training times that are reasonably close for the given durations. While much larger divergences have been recorded in previous works, testing and refitting times may be a deciding factor for model selection when under network attack.

Table 1. PyCaret model comparison results.

| Model | Accuracy | AUC | F1 | Kappa | TT (Sec) |
|---|---|---|---|---|---|
| Light Gradient Boosting Machine | 0.7787 | 0.9705 | 0.7802 | 0.7137 | 140.185 |
| Random Forest Classifier | 0.7743 | 0.9501 | 0.7767 | 0.7088 | 135.275 |
| Extra Trees Classifier | 0.7709 | 0.9452 | 0.7713 | 0.7040 | 104.255 |
| Gradient Boosting Classifier | 0.7525 | 0.9664 | 0.7664 | 0.6865 | 7778.980 |
| Decision Tree Classifier | 0.7440 | 0.8733 | 0.7451 | 0.6699 | 108.210 |
| K Neighbors Classifier | 0.7221 | 0.9003 | 0.7396 | 0.6498 | 482.320 |
| Logistic Regression | 0.6597 | 0.9568 | 0.7078 | 0.5850 | 463.130 |
| Linear Discriminant Analysis | 0.6441 | 0.9458 | 0.6910 | 0.5664 | 59.805 |
| SVM – Linear Kernel | 0.6399 | 0.0000 | 0.6870 | 0.5616 | 55.545 |
| Ridge Classifier | 0.6339 | 0.0000 | 0.6830 | 0.5550 | 48.760 |
| Naïve Bayes | 0.5044 | 0.8709 | 0.5195 | 0.3981 | 50.230 |
| Ada Boost Classifier | 0.2333 | 0.5677 | 0.2853 | 0.1287 | 214.865 |
| Quadratic Discriminant Analysis | 0.0303 | 0.4964 | 0.0074 | -0.0065 | 58.3380 |

### 3.1.1 Light Gradient Boosting Machine

LGBM showed the most improvement after a comprehensive hyper-parameter grid search. Categorical accuracy reached 81%, but with an F1 score of only 47%. This suggests that the hyperparameter tuning increased false positives and negatives from the PyCaret results. Figure 6 displays a heatmap of the resulting confusion matrix.

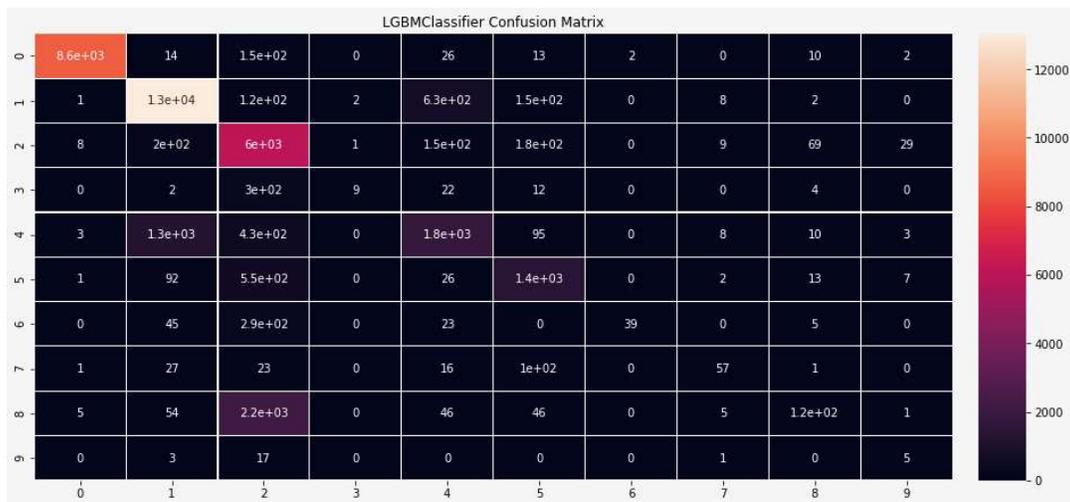

Figure 6. LGBM is able to make predictions on every class

## 3.2. Neural Networks

### 3.2.1. Convolutional

This image classification task did not result in the CNNs' proven exceptional performance. From multiple relatively small models with varying numbers of convolutional layers, to filter sizes, to combinations of normalization and pooling layers, the CNN struggled to break 80% accuracy, averaging approximately 79%. This suggests convolutional models may not be an effective method for intrusion detection data cast into an image classification problem. Figure 7 shows a typical training and validation sequence for this approach. High variance (i.e., overfitting) can be reduced with dropout and with both L1 and L2 regularization added to the fully connected (FC) backend layers. Results for CNN and FC-DNN are shown in Table 2.

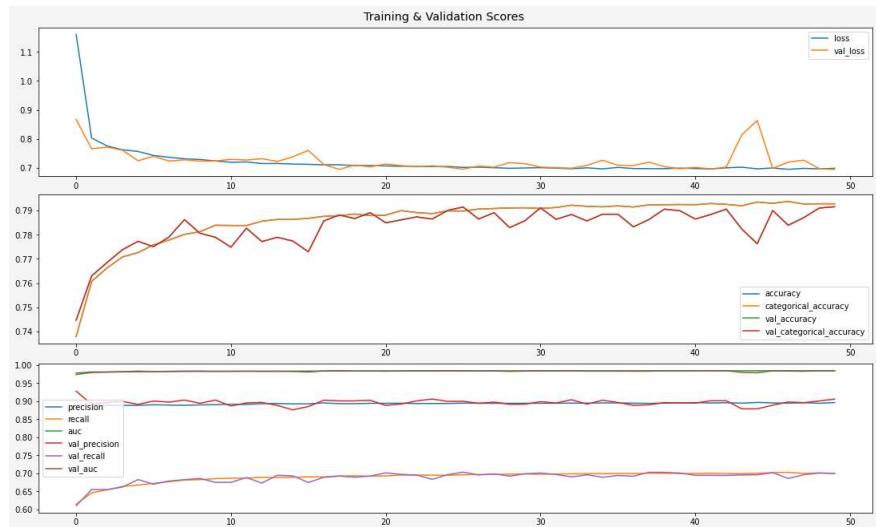

Figure 7. Convolutional neural network with four layers, including normalization and max pooling

Metrics for success are arbitrary and likely to change according to application or mission criticality. These models rely on feature detection, but the sparsity of the image arrays produces a scant few features. That lack is a factor contributing to the mediocre performance of this typically successful image classification approach. Such nuances are to be considered when deciding which metric is the deciding factor for success.

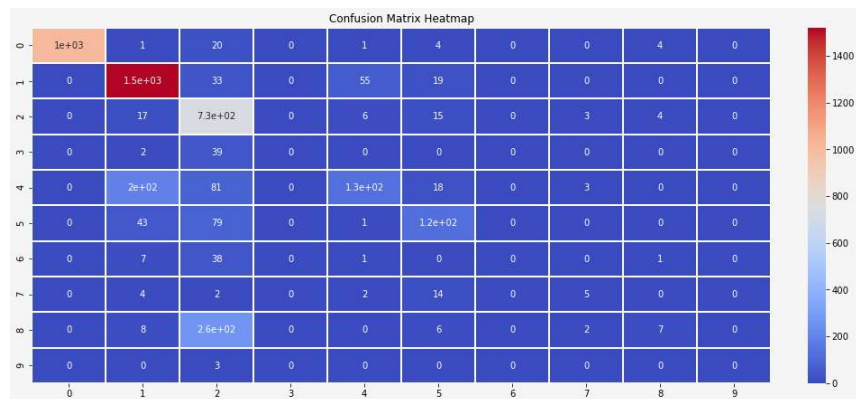

Figure 8. AlexNet makes predictions for classes 7 and 8

Most convolutional structures tested perform poorly, offering no predictions for classes 3, 6, 7, and 8. AlexNet does make predictions for classes 7 and 8 (seen in Figure 8 above) but there are only five and seven correct predictions, respectively, and those are outnumbered by the incorrect predictions. This behaviour holds even when CNN models are instructed to train for 200 epochs.

Table 2. CNN and Fully Connected Model Results

| Model | Accuracy | AUC | F1 | TT (Sec) |
|---|---|---|---|---|
| Convolutional Neural Network | 0.7969 | 0.9849 | 0.7646 | 1815.63 |
| Fully Connected Neural Network | 0.8147 | 0.9752 | 0.7767 | 1975.46 |

### 3.2.2. Fully Connected

The unexpectedly poor performance of convolutional algorithms is similarly demonstrated by the dense, fully connected (FC) perceptron architecture. 21 models with increasing hidden layer depth and varying nodes-per-layer generally achieved accuracies close to the convolutional scores; however, while CNNs and most FC NNs struggled to reach 80%, two FC models—both with six hidden layers—reach slightly over 81%. Figure 9 displays a typical training progression for this architecture.

Dense architectures show moderate improvement in predicting classes 3, 6, 7, and 8 even over AlexNet. While erroneous predictions still outnumber the correct, the F1 score is much improved over any of the tested CNNs.

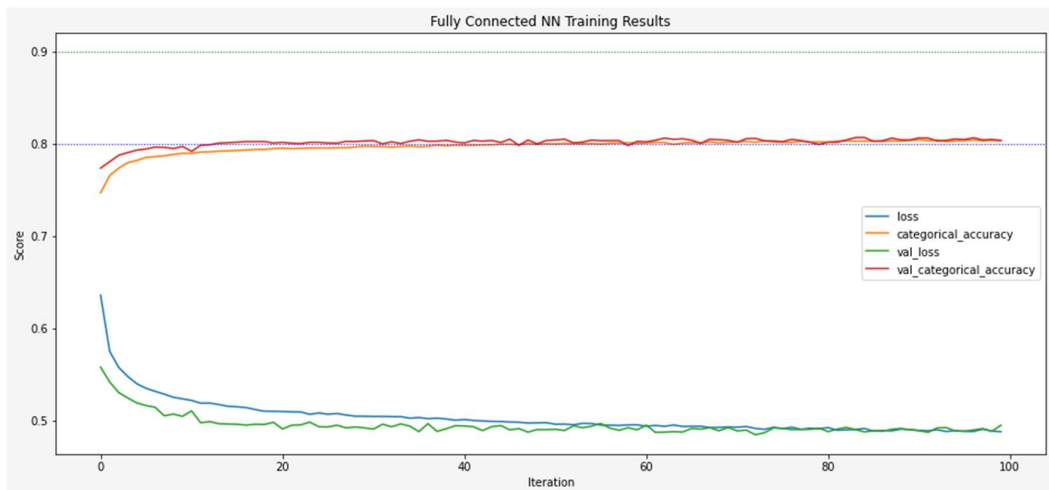

Figure 9. A dense NN achieves higher than 80% categorical accuracy with no overfitting

### 3.2.3. Graph Convolutional Network

A GCN achieved an accuracy of 72.7% and, like other models trained on this data, spent several epochs struggling to improve that value; in this instance it had reached 72% at epoch 24 of 40. Examining the adjacency graphs generated for the model, several classes are near-indistinguishable, which is a possible explanation for the poor showing. Table 3 shows results for GCN and CapsNet approaches.

### 3.2.4. CapsNet

A capsule network trained on this data performs roughly as well as the CNN, reaching an accuracy on the test data of 80%. In training, it struggles to make gains in accuracy beyond 82%. The capsule network performs well only against some classes; it is possible this is due to the uneven distribution of classes in the dataset.

Table 3. CapsNet and GCN Results

| Model   | Accuracy | Precision | Recall | F1   | TT (Sec) |
|---------|----------|-----------|--------|------|----------|
| CapsNet | 0.80     | 0.79      | 0.80   | 0.78 | -        |
| GCN     | 0.72     | -         | -      |      | 1975.46  |

## 5. CONCLUSIONS

An optimized Light Gradient Boosting Machine (LGBM) yields the best performance with an accuracy of 0.8055 but a poor F1 score of 0.47. It can make predictions on every class whereas many other algorithms, like AlexNet, cannot. By using dropout layers and L1/L2 regularization where applicable to eliminate overfitting, a fully connected, dense neural network performs as well as LGBM and with the same score and a higher F1. We find LGBM to be superior, however, due to its ability to make predictions on every class and with greater accuracy than the dense neural network.

## ACKNOWLEDGEMENTS

The authors would like to thank the PeopleTec, Inc. Technical Fellows for facilitating this research.

## Authors

Erik Larsen, M.S. is a senior data scientist specialized in quantum physics and deep learning. Prior research includes electronic band structure calculations with density functional theory. He completed both M.S. and B.S. in Physics at the University of North Texas, and a B.S. in Professional Aeronautics from Embry-Riddle Aeronautical University while serving as an aviator in the U.S. Army.

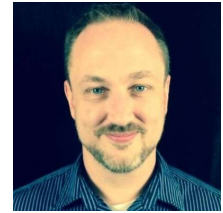

Korey MacVittie, M.S. is a data scientist specialized in machine learning. Prior research includes identifying undervalued players in sports drafting. He completed his M.S. at Southern Methodist University, and a B.S. in Computer Science and B.S. in Philosophy from University of Wisconsin Green Bay.

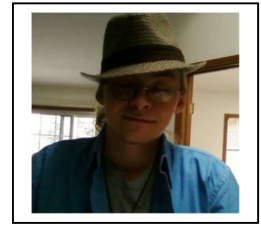

John R. Lilly III is an applied machine learning research scientist specializing in quantitative analysis, algorithm development, and implementation for embedded systems. Prior research is in token economics and the deployment of crypto assets across multiple enterprise blockchain protocols. He has been trained in statistics by Cornell University. Formerly infantry, he currently serves as an intelligence advisor for the U.S. Army's Security Force Assistance Brigade.

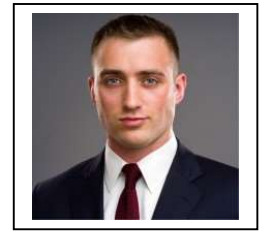